# Measuring Online Hate on 4chan using Pre-trained Deep Learning Models

Adrian Bermudez-Villalva, Maryam Mehrnezhad, and Ehsan Toreini

*Abstract*—Online hate speech can harmfully impact individuals and groups, specifically on non-moderated platforms such as 4chan where users can post anonymous content. This work focuses on analysing and measuring the prevalence of online hate on 4chan's politically incorrect board (/pol/) using state-of-the-art Natural Language Processing (NLP) models, specifically transformer-based models such as RoBERTa and Detoxify. By leveraging these advanced models, we provide an in-depth analysis of hate speech dynamics and quantify the extent of online hate non-moderated platforms. The study advances understanding through multi-class classification of hate speech (racism, sexism, religion, etc.), while also incorporating the classification of toxic content (e.g., identity attacks and threats) and a further topic modelling analysis. The results show that 11.20% of this dataset is identified as containing hate in different categories. These evaluations show that online hate is manifested in various forms, confirming the complicated and volatile nature of detection in the wild.

*Index Terms*—Hate speech, machine learning, natural language processing (NLP), online hate, toxicity analysis.

## I. INTRODUCTION

THE SPREAD of hate speech on online platforms has become a serious problem in our society. As digital communication becomes ubiquitous, platforms like 4chan, known for their anonymity and minimal moderation, have become hotspots for this harmful behaviour. This is particularly evident on its politically incorrect board, /pol/, a notorious board dedicated to discussing politics and current events, often associated with hate speech, extremist content, and conspiracy theories [1]. The anonymity provided by these platforms often encourages users to express extreme ideologies [2]. This issue raises significant concerns about the impact on at-risk and vulnerable groups as it can cause real-world harm, including psychological trauma. Therefore, a systematic approach is needed to measure and understand the prevalence and forms of online hate. In addition to hate speech,

Received 28 August 2024; revised 23 December 2024, 10 February 2025, and 6 March 2025; accepted 6 March 2025. This work is supported by the UK Research and Innovation (UKRI), through the Strategic Priority Fund as part of the Protecting Citizens Online programme (AGENCY: Assuring Citizen Agency in a World with Complex Online Harms, EP/W032481/2). *(Corresponding author: Adrian Bermudez-Villalva).*
Adrian Bermudez-Villalva and Maryam Mehrnezhad are with the Information Security Department, Royal Holloway University of London, TW20 0EX Egham, U.K. (e-mail: adriano.bermudezvillalva@rhul.ac.uk; Maryam.Mehrnezhad@rhul.ac.uk).
Ehsan Toreini is with the School of Computing, Surrey University, GU2 7XH Guildford, U.K. (e-mail: e.toreini@surrey.ac.uk).
Digital Object Identifier 10.1109/TTS.2025.3549931

toxic content worsens the environment on these platforms [3]. This content is often linked to hate speech, making it crucial to address both issues together to understand the negative dynamics in digital spaces.

Machine Learning (ML), specifically Natural Language Processing (NLP), presents a promising avenue for analysing such speech, offering tools to sift through vast amounts of data to identify patterns and instances of online hate and toxic content. Recent advancements in the field, have led to the development of sophisticated models capable of understanding nuances in language and context, which are crucial for accurately identifying harmful content in its various forms. Deep learning, particularly transformer-based models, has significantly enhanced these capabilities, enabling more context-aware and accurate classification of online hate speech.

Research shows that online hate is a complex and prevalent issue, manifesting in various forms targeted at different groups based on race, gender, religion, and more [4]. Previous studies have applied NLP techniques to detect and analyse online hate across mainstream platforms like Twitter and Facebook [5]. However, less attention has been paid to platforms like 4chan, where the anonymous and ephemeral nature of the content poses unique challenges for the detection and analysis of online hate, especially due to its user base, linked to alt-right communities considered hateful, racist, and misogynistic [1].

The academic community has recognised the utility of pre-trained models such as BERT and GPT for their ability to classify online hate [6]. These tools offer a way to process large datasets without the need for model training, making them ideal for studies focusing on dynamic and extensive data derived from social media platforms [35]. Despite these advancements, a significant research gap remains in understanding the extent of online hate, as most studies have either focused on more mainstream platforms, concentrated on tuning, training, and improving model performance, or on testing and comparing models, rather than applying these models to extract meaningful patterns from the data to perform analysis that provides insights into real-world behaviours. Additionally, many of these studies have not differentiated between the types of hate speech and their targeted groups in the context of less moderated environments (see Table I).

The primary objective of this study is to measure and analyse the prevalence of online hate on 4chan's /pol/. By utilising state-of-the-art transformer-based models, this study aims to fill the existing research gap by conducting an examination of how hate speech manifests in an unregulated environment. Instead of evaluating or comparing models, we focus on applying well-performing models from prior research

TABLE I
RELATED WORK ON ONLINE HATE DETECTION

| Year | Type of Hate | Platform | Size of Dataset | ML/AI Technique | Source |
|------|--------------|----------|-----------------|-----------------|--------|
| 2016 | Racism | Twitter | 136K | Character n-grams | [8] |
| 2017 | General | Twitter | 85M | Logistic Regression, Naïve Bayes, Decision Trees, Random Forest, SVM | [9] |
| 2018 | Religion | 4chan, Gab | 100M | word2vec | [10] |
| 2019 | Sexism | Twitter | 12K | Support Vector Machines | [11] |
| 2020 | General | YouTube, Reddit, Wikipedia, Twitter | 197K | Logistic Regression, Naïve Bayes, SVM, XGBoost, ANN | [5] |
| 2021 | Sexual Orientation | Twitter | 24K | Naïve Bayes, Logistic Regression, Stochastic Gradient Descent, SVM, ANN | [12] |
| 2023 | Multiclass | Twitter | 83K | Transformed-based models | [29] |
| 2023 | Sexism | Twitter | 8M | Transformed-based models | [13] |
| 2024 | Multiclass | 4chan | 0.5M | Transformed-based models | Our work |

to real, in-the-wild data to understand the prevalence and dynamics of hate on 4chan. Our research questions include:

**Research Questions: RQ1:** What is the prevalence of different types of hate speech on 4chan's /pol/ board? **RQ2:** How does the online hate speech identified vary in terms of its toxic context? **RQ3:** What are the predominant topics discussed within the contexts of hate speech against vulnerable groups on 4chan?

This study advances the academic understanding of online hate dynamics on a notoriously unregulated platform through improved analysis methodologies and aids in developing more targeted interventions by policymakers and platform developers. The contributions of this research are as follows:

- This study provides an **empirical assessment** of hate speech prevalence on the /pol/ board of 4chan. It advances our understanding of how different types of hate are distributed in an anonymous and non-moderated context.
- This work utilises the state-of-the-art NLP models, specifically **fine-tuned versions of transformer-based models**, to detect and quantify instances of online hate targeting various communities. This application in a real-world setting marks a significant step forward in moving from theoretical model testing to practical measurement of online hate.
- The research extends beyond online hate detection by also **classifying toxic content**, including general toxicity, identity attacks, and threats. This analysis provides a more comprehensive understanding of the hostile environment on 4chan.
- The study further contributes by applying **topic modelling** to uncover key themes within the most prominent hate speech categories identified (racism, sexism, sexual orientation, and religious hate). This analysis deepens the understanding of the contextual factors that drive online hate.
- This study contributes to the body of knowledge by building and releasing a **dataset of 0.5 million 4chan posts** [40], focusing on the /pol/ board. This dataset enables researchers to analyse hate speech dynamics, evaluate machine learning models, and study the societal impact of unmoderated online platforms.

By offering empirical evidence and advanced analytical tools, this study informs the development of more effective content moderation policies and platform governance strategies.

## II. BACKGROUND AND RELATED WORK

The rise of digital platforms and the internet has led to the prevalence of online hate and toxic content across various social media, websites, and forums, expressed through images, text, and videos. Mainstream platforms like Facebook, Instagram, and Twitter are significant sources of online hate [6], [11]. However, this issue is even more pronounced in obscure communities such as 4chan, where the lack of moderation fosters the spread of hate speech.

### A. Definition of Online Hate

The phenomenon of online hate has its roots in the broader context of hate speech, which has existed long before the advent of digital platforms. However, the widespread use of the internet and social media has amplified the visibility and impact of online hate. Initially reflecting offline prejudices and discrimination against vulnerable groups, online hate has evolved due to the anonymity and expansive reach of digital platforms, becoming a more complex phenomenon. It now includes a broad spectrum of expressions, ranging from aggressive remarks to subtler forms of bias and exclusion [14]. Determining the boundary between hate speech and offensive language presents a challenge, as no formal definition exists. However, there is an agreement among scholars and practitioners that hate speech targets socially marginalised groups [9], [14].

Online hate then can be broadly defined as the use of digital platforms to disseminate content that denigrates, intimidates, or incites violence against individuals or groups based on attributes such as gender, race, ethnicity, religion, sexual orientation, or other identity markers. This definition aligns with the European Union's definition of hate speech, emphasising public incitement of violence or hatred based on personal attributes [15]. It is important to note that this definition emphasises that online hate targets collective identities rather than individuals, setting it apart from other forms of online harassment or bullying [6].

### B. Identifying Online Hate

Therefore, the identification of online hate involves categorising language that is not only derogatory but also specifically targets marginalised groups or individuals. For instance, hate categories might include racism, sexism, homophobia, religious hate, and disability hate. Over the years, online platforms have developed their own categorisation of hate speech aimed at moderating user-generated content [14], with the following distinct characteristics and targets: **Racial Hate**: Targets individuals or groups based on racial or ethnic backgrounds, often including derogatory language, and incitement to violence. **Religious Hate**: Directed against specific religious communities and involves criticism of religious beliefs and attacks on the followers themselves. **Gender-Based Hate**: Encompasses misogyny, sexism, and attacks on individuals based on their gender or those who do not fit into traditional gender roles. It often manifests in threats, sexual harassment, and demeaning stereotypes. **Sexual Orientation Hate**: Targets LGBTQ+ individuals, attacking their sexual orientation, gender identity, or expression. **Disability Hate**: Prejudice and discrimination against individuals with physical or mental disabilities, often expressed through mockery or exclusion. **Political or Ideological Hate**: Involves attacks based on political beliefs or affiliations, often seen in polarised political environments.

*C. Identifying Toxic Content*

Building on the identification of online hate, it is important to consider toxic content, a broader category that includes harassment, threats, and offensive language. While online hate specifically targets individuals or groups based on distinctive characteristics, toxic content contributes to general negativity, degrading the quality of online interactions and potentially creating an environment conducive to online hate [13], [16]. Analysing toxic content alongside online hate is essential for a comprehensive understanding of hate dynamics.

*D. 4chan*

4chan serves as a pertinent example of an online platform where online hate proliferates. 4chan is an imageboard website composed of user-generated content dedicated to a specific interest, such as video games, music, literature, and politics. The site is known for its minimal moderation and the anonymity it affords its users, factors that have contributed to its reputation as a breeding ground for controversial and often extremist content [1].

4chan follows an imageboard structure, where discussions are organised into boards, threads, original posts (OP), and replies. Boards host a limited number of active threads at a given time, and as new threads are created, older threads are pushed down and eventually archived [17]. Threads with higher activity, measured by the number of posts, are prioritised and kept at the top of the board. However, 4chan imposes a maximum limit of 300 times a post can be bumped to the top, after that, the thread is moved to the archive [18].

The /pol/ board, short for "Politically Incorrect" is one of 4chan's most infamous sections. It is characterised by discussions on political and social issues, often delving into hate speech, conspiracy theories, and radical ideologies. Figure 1 shows an example of a 4chan misogynist post. The content on /pol/ has been linked to various online and offline incidents of hate and violence, showcasing the board's influence and the potential dangers of unmoderated online spaces. The anonymity and lack of accountability on /pol/ facilitate a culture where extreme views are expressed freely, making it a significant challenge for researchers aiming to detect and analyse online hate.

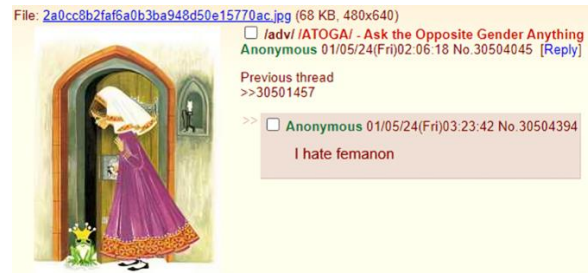

Fig. 1. An example of a 4chan post.

Some studies have provided a detailed examination of the content and culture within the /pol/ board, highlighting the role of anonymity in the spread of controversial and hateful content [17], [18]. For instance, previous research has conducted basic content analysis, identifying key discussion topics and levels of toxicity [19]. This prior work focused on preliminary content analysis without utilising NLP models for deeper insights. In contrast, the present study employs advanced models to provide a detailed analysis of online hate on /pol/ to uncover complex patterns, offering new insights into the dynamics of hate speech on minimally moderated platforms.

*E. Online Hate Detection*

Early research in this domain primarily relied on lexicon-based and rule-based approaches, which utilised predefined lists of offensive terms and syntactic patterns to detect hate speech. While these methods were straightforward and interpretable, they struggled to account for the dynamic and contextual nature of hate speech, leading to significant limitations in accuracy and adaptability [20].

With the advent of machine learning, more sophisticated approaches were introduced, leveraging models such as Support Vector Machines (SVMs) and logistic regression to classify text into hate and non-hate categories. These models used engineered features like word frequencies and n-grams to capture linguistic patterns, offering improved accuracy compared to earlier methods. However, their reliance on manually crafted features limited their scalability and ability to adapt to the rapidly changing language and context of online hate [4], [6], [7], [21].

The advent of deep learning brought significant advancements in this field. Particularly, the development of transformed-based models [24]. For instance, BERT (Bidirectional Encoder Representations from Transformers) introduced a new paradigm for NLP tasks, including online hate detection [25]. BERT was pre-trained and developed by

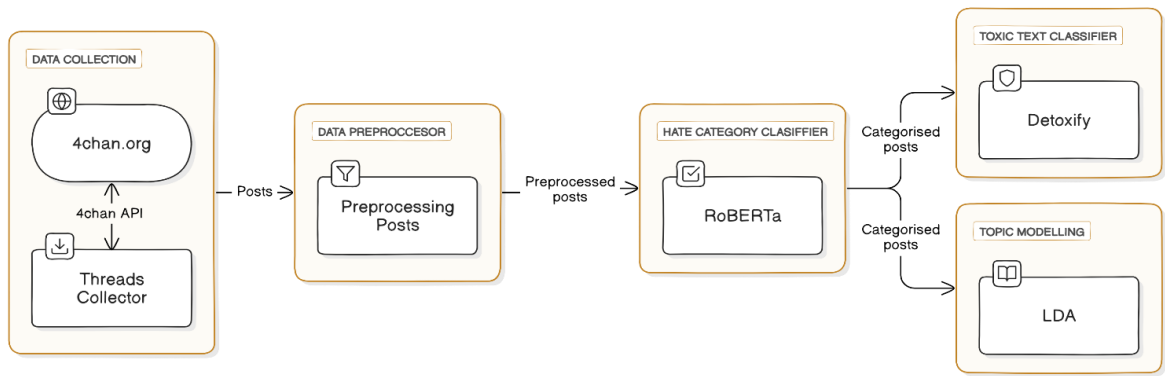

Fig. 2. Overview of the methodology used to measure online hate on 4chan.

Google based on the knowledge extracted from text (vectors) using the surrounding text to establish the context. These models, pre-trained on large corpora of text, could be fine-tuned on specific tasks, achieving state-of-the-art results due to their deep contextual understanding [5]. This was further enhanced by the introduction of RoBERTa [26], an optimised version of BERT, which improved training procedures and model architecture for even better performance.

Fine-tuned versions of RoBERTa have been used to improve hate speech classification [27], [28]. For instance, the RoBERTa Base Hate Multiclass model has been trained on different multi-platform datasets to identify online hate targeting different communities [29]. Similarly, toxicity classification models based on BERT have been trained to classify toxic language [3], [16].

Academic research into online hate speech detection has extensively employed a range of techniques focusing on the binary classification of content as hateful or not hateful and categorise expressions of hate targeting various at-risk groups (Table I). This body of research includes studies that target misogynistic and sexist content aimed at women and girls [11], [13], [22], racism [8], religious hatred [10], attacks based on sexual orientation [12] and disabled individuals [23].

Most of these studies have not explicitly analysed the extent of online hate but have compared NLP techniques, models and datasets in terms of precision, accuracy and data quality. There is a lack of studies applying these techniques and models to detect and quantify online hate in the wild. This study encompasses a measurement of several vulnerable groups, aiming to provide a comprehensive analysis of online hate across different communities. Doing so bridges the gap between theoretical research and practical application, providing insights into the prevalence of online hate on 4chan.

Different to previous works using transformed-based models [13], this study uses the RoBERTa Base Hate Multiclass to perform a multiclass classification of hate against different vulnerable groups. By applying advanced NLP techniques to real-world data from 4chan, this work aims to measure the prevalence of online hate, uncover patterns, and provide insights into the dynamics of hate speech on minimally moderated platforms.

III. METHODOLOGY

Utilising transformer-based multi-class models trained across several datasets, this research seeks to apply robust NLP tools to explore online hate in a relatively uncharted territory. The exploratory analysis aims to provide a comprehensive understanding of the prevalence and characteristics of online hate on 4chan's /pol/ board and the categorisation of multidimensional hate against different communities. It also enhances our understanding of how such speech proliferates in less regulated environments. In addition, by conducting a topic modelling analysis, this research further seeks to uncover underlying themes and discourses within large text corpora, offering insights into the contextual factors that perpetuate hate speech online.

In this research, the focus is on using pre-trained NLP models to quantify online hate on 4chan. In this section, the methodology employed for collecting and analysing a large corpus of 4chan posts is presented. As shown in Figure 2, the 4chan API is used to gather threads and posts, which are then preprocessed. Subsequently, the prevalence of online hate and toxic content on 4chan is measured using state-of-the-art pre-trained models.

*A. Data Collection*

The data collection methodology for this study follows the approach detailed in [19]. The collection process started in May 2024, using the JSON API (github.com/4chan/4chan-API) provided by 4chan to crawl the /pol/ board. The 4chan API provides endpoints that allow researchers to retrieve content from specific boards in a structured JSON format. Due to the ephemeral nature of the content on 4chan, the thread catalogue of /pol/ is retrieved every 5 minutes, using the board's API endpoint (*https://a.4cdn.org/pol/catalog.json*), to capture the full and final contents of all threads. Each time the board is retrieved, the list of active threads is compared with those recorded previously. Once a thread is identified as no longer active, a complete copy of that thread is obtained from 4chan's archive. For each thread, only the textual content of the posts is collected for the purpose of analysis. No additional data, such as user information, timestamps, or any other metadata, is retrieved, as these are not required for the scope of this study. The data released along with this paper aligns with the FAIR guiding principles for scientific data (go-fair.org/fair-principles/).

It is important to note that 4chan operates on a highly anonymous basis. Users do not register accounts, and therefore do not have usernames or identifiable profiles linked to their activity. Instead, they have the option to use "tripcodes" which are unique hashes based on a password provided by the user, but even these are optional and do not provide any consistent identifier that could be traced back to real-world identities [34]. Given the anonymous nature of 4chan, posts generally do not contain structured personally identifiable information (PII) that could be used to identify a specific individual. Even if users provide such information, it is not collected for this study.

*B. Preprocessing*

The preprocessing stage for ML and NLP models is crucial to preparing data in the format and structure required by specific models. Since NLP pre-trained models can handle traditional preprocessing steps such as tokenisation, lemmatisation, and feature extraction, a simplified preprocessing task is applied to prepare the data for the chosen models. Given that 4chan posts often include highly informal language, slang, misspellings, abbreviations, and non-standard language use, the following steps have been designed to address these characteristics effectively:

**Cleaning:** In this stage, first, the HTML tags, URLs, and unnecessary metadata are removed from the dataset. This is typically performed using regular expressions or HTML parsing libraries. Next, the greater-than symbol (>) is used in 4chan to indicate quoted text or may appear as part of HTML code or URLs. Depending on its function in the text, this symbol is processed to enhance the efficiency of the task.

| Example Post | Cleaned Text |
|---|---|
| `<p>>OP is a f****t. Check out this meme: <a href="http://example.com">link</a></p>` | *OP is a f****t. Check out this meme:* |

**Text Normalisation:** In this phase, we follow three steps: (i) **Lowercasing:** where all text data is converted to lowercase to ensure uniformity, (ii) **Removing non-ASCII Characters:** where posts containing non-ASCII characters are normalised or removed, and (iii) **Handling Slang and Abbreviations:** where a dictionary of common 4chan slang and abbreviations is developed to replace or expand them into their full forms.

| Example Post | Normalised Text |
|---|---|
| *">Anon thinks all women are THOTs. KEK!"* | *"anonymous thinks all women are that hoe over there. laugh out loud!"* |

*C. Hate Category Extraction*

Transformer-based models, such as BERT and RoBERTa, have been recognised for their effectiveness in identifying hate speech targeting various groups [27]. These models, when fine-tuned for multi-class classification, have demonstrated superior performance compared to traditional machine and deep learning models. Additionally, their effectiveness is further enhanced when trained on data from multiple platforms, facilitating the task of categorising multidimensional hate against diverse communities, especially in unstructured online environments like 4chan [5].

For the categorisation of hate posts, the *RoBERTa Base Hate Multiclass* model is selected [29]. This open-source model, trained on a diverse collection of hate speech detection datasets encompassing 83,000 tweets from various platforms, utilizes self-attention mechanisms to assign contextual weights to tokens (words) based on their relevance within a post. To categorise posts, RoBERTa breaks each post into smaller parts (tokens) and processes them to understand the relationships between words. It then assigns a category (such as racism or sexism) based on which category best matches the content. The model uses a scoring system to decide which label fits best, and the category with the highest score is chosen [26].

In the classification process, RoBERTa was trained and fine-tuned to classify posts into the following categories: *sexism, racism, disability hate, hate based on sexual orientation, religious hate, other types of hate, and non-hate speech* [39]. For this study, each post is assigned to a single category based on the most probable output from the model. While hate speech can exhibit overlapping characteristics, only the dominant category predicted by the model is considered. This ensures that each post is uniquely classified, maintaining clarity and consistency in the analysis.

The RoBERTa Base Hate Multiclass model achieved an accuracy of 0.9419, a macro-F1 score of 0.5752, and a weighted-F1 score of 0.9390 [39]. These metrics demonstrate the model's strong overall performance, particularly in terms of accuracy and weighted-F1, while highlighting the challenges of achieving high macro-F1 due to the varying difficulty of detecting certain hate categories.

*D. Toxic Content Extraction*

Toxic content classification refers to automatically identifying and categorising offensive or harmful language in text. Detoxify, an open-source model developed by Unitary AI, is utilised for this purpose [30]. This model, which is a finely tuned version of BERT, has been trained on the dataset from the Jigsaw Toxic Comment Classification Challenge [31]. Detoxify is chosen for this study due to its competitive performance and the flexibility it offers for local installation and operation, avoiding the constraints typically associated with web API usage found in other classifiers. In this task, Detoxify is applied to categorise 4chan posts into one of the following categories: **Toxicity**: General offensive or unpleasant comments that could make someone leave a discussion. **Severe Toxicity**: More aggressive and hateful comments. **Obscenity**: Comments containing vulgar or inappropriate language. **Threat**: Comments intending to inflict harm or violence. **Insult**: Comments that are demeaning or rude. **Identity Attack**: Comments that attack a person or a group based on attributes such as race, religion, ethnic origin, sexual orientation, disability, or gender. **Sexual Explicit**: Unwanted or offensive sexual advances or remarks.

*E. Topic Modelling*

Topic modelling analysis is also applied to systematically uncover frequent themes and subtopics within each category of online hate on 4chan, enabling a deeper understanding of the contextual framework within which online hate is propagated. This analysis is performed using Latent Dirichlet Allocation (LDA), a statistical model that discovers abstract topics within a collection of documents [32]. LDA assumes documents are mixtures of topics, where each topic is a distribution over words.

TABLE II
SUMMARY OF DATA COLLECTION

| Description | Number of Posts |
|---|---|
| Valid Posts After Preprocessing | 476,759 |
| Empty Records Removed | 23,241 |
| Total Posts Collected | 500,000 |

*F. Ethical Considerations*

This work received ethical approval from the authors' institution. Data for this study is exclusively collected from publicly available posts on 4chan. Only the textual content of these posts is gathered with no additional data is extracted. IDs of the individuals who post these items are not collected. Users on 4chan do not register accounts or use identifiable usernames. Instead, they may optionally use "tripcodes" which are not linked to real-world identities. We don't collect any of such data. Consequently, the dataset does not contain personally identifiable information (PII) as typically defined under GDPR. Only the raw textual content of the posts was included for the publication of the dataset. [40].

IV. RESULTS

This section outlines the results derived from the analysis of the dataset collected via the 4chan API.

*A. Dataset Overview*

As shown in Table II, a total of 500,000 posts were initially collected, with preprocessing activities resulting in 23,241 records being identified as empty and subsequently removed from our analysis. The valid posts amounted to 476,759.

*B. Online Hate in General*

The RoBERTa Base Hate Multiclass model is employed to classify each post into either a specific category of hate (sexism, racism, disability hate, hate based on sexual orientation, religious hate, or other types of hate) or into a *non-hate* category [29]. To provide a general characterisation of hate within the dataset, all specific hate categories are aggregated under a single 'Hate' label, which was then compared against the 'Non_Hate' category.

Out of the total valid posts, 52,948 posts (11.20%) are identified as containing hate, while the remaining 419,765 posts (88.80%) are categorised as not containing hate. These findings underscore the presence of significant online hate on 4chan, suggesting that a notable proportion of the discourse encompasses hate speech.

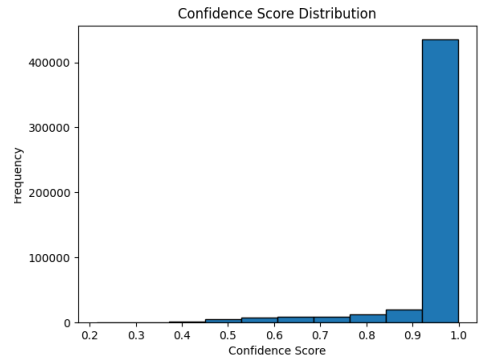

Fig. 3. Confidence score distribution of model predictions.

TABLE III
CATEGORIES OF ONLINE HATE

| Category | Percentage |
|---|---|
| Racism | 35.92% |
| Religion | 23.36% |
| Sexual Orientation | 16.49% |
| Sexism | 12.01% |
| Other | 10.49% |
| Disability | 1.73% |

To evaluate the model's confidence in its predictions, the confidence score distribution is analysed across the dataset [36]. The confidence score represents the probability assigned by the model to the predicted hate speech category for each post. Figure 3 illustrates that most of the predictions exhibit high confidence scores (above 0.9). This suggests that the model is confident in its classifications for most instances. A small proportion of predictions fall below a confidence threshold of 0.9, indicating cases where the model was less certain. In general, the resulting distribution highlights the model's strong prediction performance.

*C. Hate Categorisation Analysis*

The breakdown of specific hate categories within the data classified as hate can be seen in Table III. This categorisation reflects the distribution of hate speech, with racism identified as the most prevalent type, accounting for more than a third of the hateful messages (35.93%). Other substantial categories include religion (23.36%), sexual orientation (16.50%), and sexism (12.01%). The category labelled 'Other', accounts for 10.48% of the messages, encompassing forms of hate speech that do not specifically fall into the conventional categories but are equally concerning due to their impact. For instance, derogatory remarks aimed at individuals for being unemployed. Although hate speech related to disability is less frequent, at 1.73%, it remains a critical area of concern due to its potential impact on vulnerable populations.

In the absence of true labels for the dataset, an **intrinsic evaluation** was conducted to assess the behaviour and performance of the model [37], [38]. Intrinsic evaluation focuses on analysing the internal properties of the model's predictions rather than comparing them to ground truth labels. It assesses key aspects such as prediction distribution, diversity, and consistency to gain insights into the model's behaviour. To make this evaluation robust, we employed k-fold cross-validation, a technique that splits the dataset into k parts and evaluates the model on each subset to analyse its performance across different splits. Although k-fold cross-validation typically involves training a model on k-1 folds and testing it on the remaining fold, in this study, no additional training was performed. The pre-trained nature of the model allowed us to focus solely on its prediction behaviour and stability across folds without introducing variability from re-training.

TABLE IV
TOXIC CONTENT ANALYSIS

|  | Toxicity | Identity Attack | Obscene | Insult | Sexual Explicit | Threat |
|---|---|---|---|---|---|---|
| Racism | 97.56% | 2.11% | 0.26% | 0.07% | 0.01% |  |
| Religion | 96.90% | 3.30% | 0.08% | 0.02% |  |  |
| Sexual Orientation | 99.55% | 0.09% | 0.21% | 0.11% | 0.02% | 0.01% |
| Sexism | 97.51% | 0.17% | 1.01% | 0.02% | 1.29% |  |
| Other | 88.50% | 0.56% | 8.06% | 2.83% | 0.04% | 0.02% |
| Disability | 98.36% | 0.11% | 0.98% | 0.55% |  |  |

TABLE V
TOPIC MODELLING ANALYSIS

| Racism |
|---|
| .008*"fuck"+.008*"chinese"+.007*"like"+.006*"kikes"+.006*"sandniggers" |
| .028*"fuck"+.022*"stupid"+.012*"zigger"+.012*"shit"+.011*"niggers" |
| .078*"nigger"+.016*"fucking"+.016*"white"+.014*"fuck"+.012*"like" |
| .022*"china"+.016*"ignorant"+.004*"us"+.004*"country"+.004*"send" |
| .051*"niggers"+.013*"like"+.011*"negroid"+.011*"would"+.010*"people" |

| Religion |
|---|
| .106*"arab"+.014*"muslims"+.010*"people"+.009*"hate" .008*"terrorist" |
| .041*"jew"+.014*"like"+.008*"filthy"+.008*"fucking"+.007*"ironic" |
| .031*"fuck"+.023*"jew"+.014*"shit"+.013*"jewish"+.013*"fucking" |
| .059*"dirty"+.035*"jews"+.013*"jewish"+.007*"kike"+.006*"rats" |
| .043*"jews"+.021*"greedy"+.007*"get"+.006*"would"+.006*"like" |

| Sexual Orientation |
|---|
| .038*"faggots"+.035*"gay"+.012*"shit"+.011*"fucking"+.008*"fags" |
| .028*"homosexual"+.025*"faggots"+.015*"like"+.014*"fucking"+.009*"fag" |
| .084*"homosexual"+.017*"fuck"+.014*"fucking"+.010*"like"+.010*"get" |
| .018*"faggy"+.014*"fags"+.005*"liberal"+.004*"got"+.004*"nothing" |
| .018*"newfag"+.016*"like"+.016*"pussy"+.009*"fucking"+.009*"fags" |

| Sexism |
|---|
| .023*"whore"+.014*"women"+.014*"like"+.009*"would"+.008*"whores" |
| .016*"slut"+.014*"fuck"+.012*"get"+.012*"women"+.010*"fucking" |
| .045*"women"+.014*"men"+.013*"fuck"+.010*"hate"+.008*"want" |
| .018*"cunt"+.017*"bitch"+.014*"man"+.012*"whore"+.012*"weak" |
| .019*"bitch"+.017*"like"+.014*"women"+.012*"pussy"+.010*"fuck" |

The intrinsic evaluation employed 5-fold cross-validation, dividing the dataset into five equal parts. The results of the evaluation indicate that the model consistently produces the same output for each record across runs and folds. This stability demonstrates a highly reproducible prediction process, which is critical for reliable classification in practical applications. However, while stability and consistency are desirable, the diversity of predictions is equally important for capturing the complexity of hate speech on the platform. The overall prediction distribution (Table III) indicates that the model effectively identifies various categories of hate speech. Although categories like disability are less frequently predicted, this aligns with the lower prevalence of such content in the dataset.

The intrinsic evaluation supports the notion that the model captures hate categories with high consistency, as evidenced by a consistency score of 100% across runs and folds.

*D. Toxic Content Analysis*

The application of the Detoxify model to classify the nature of hate speech on 4chan reveals significant insights into the distribution of toxic content across various categories. Table IV shows the aggregated Detoxify toxicity percentages across posts classified into hate categories by RoBERTa, as we discussed in the Methodology section. The predominance of general toxicity in nearly all categories, particularly in posts related to sexual orientation (99.55%), racism (97.56%), and sexism (97.51%), suggests that these forms of hate speech are often expressed through direct and overtly offensive language.

The relatively higher occurrence of identity attacks in discussions related to religion (3.30%) compared to other categories could be indicative of the specific nature of religious discourse, which often intersects with deeply held beliefs and identities, making it a more likely target for identity-based attacks.

Obscenity shows a notable variance, with the 'Other' category registering a significantly higher rate (8.06%). This could be due to the miscellaneous nature of this category, which may include a variety of topics that do not fit neatly into predefined categories, potentially leading to a broader spectrum of vulgar expressions.

Insults and threats, while generally low across all categories, show a slightly higher presence in the 'Other' category (insults at 2.83% and threats at 0.02%). This again might be attributed to the diverse and possibly more contentious nature of the topics grouped under 'Other', which could provoke more personal confrontations.

Sexually explicit content is most prevalent in the sexism category (1.29%), which is consistent with the nature of sexist discourse that often includes unwanted or derogatory sexual remarks targeted at specific genders.

*E. Topic Modelling Analysis*

The analysis examines the most prominent topics discussed within the hate categories in the dataset. Only the most dominant hate speech categories, as identified in our previous results, are selected for this analysis (Table III). This is crucial for understanding not only the presence of hate speech but also the specific contexts and themes. By examining these categories, we can uncover underlying patterns that might not be evident through simple content categorisation. The topics were not manually selected or adjusted; instead, the topics were automatically generated based on the algorithm's output as it identified patterns within the data. The decision to set the model for five topics was made as a practical choice to maintain simplicity while exploring the most significant trends in the dataset.

Table V presents the findings from the topic modelling analysis using Latent Dirichlet Allocation (LDA). Each row in the table represents a topic, with the most significant words for that topic listed alongside their respective weights. These weights indicate the importance of each word within the topic, with higher values signifying greater relevance. For instance, in the Racism category, the word "nigger" appears frequently with high weights across multiple topics. Similarly, in the Religion category, terms like "jew" and 'muslims" are prominent.

Topics classified under **racism** reveal discussions that often combine explicit racial slurs with aggressive language, alongside mentions of broader societal and political issues. The consistent presence of highly derogatory terms across all topics emphasises the severity and blatant nature of the racist language used. This suggests that racist discourse is not only confined to isolated incidents but is part of broader discussions that often intersect with major societal concerns.

The analysis of topics related to **religion** shows a significant amount of hostile language directed predominantly towards Jewish and Muslim communities. This reflects a considerable level of religious intolerance, with frequent usage of specific slurs and aggressive language. These discussions are frequently interwoven with political and social commentary, indicating that religious hate speech on the platform often overlaps with broader geopolitical and cultural debates.

For the category of **sexual orientation**, the topics reveal a pronounced use of derogatory slurs and hostile language aimed at individuals based on their sexual preferences. The pervasive use of offensive terms underscores the harshness of hate speech related to sexual orientation. This suggests a hostile environment for LGBTQ+ individuals, where negative stereotypes and aggressive rhetoric are commonplace.

The topics identified within the **sexism** category demonstrate a prevalent use of aggressive and abusive language. The repetitive occurrence of words like 'bitch', 'whore', and 'cunt' across various topics highlights these as common terms employed in sexist discourse on the platform. Such findings point to a deeply ingrained culture of misogyny and objectification of women within the community, where derogatory language is normalised and frequently used.

The topic modelling results provide a stark depiction of the thematic structures of hate speech across different categories on 4chan. The consistent theme across all categories is the use of intensely derogatory and aggressive language, which not only vilifies individuals based on their identity but also perpetuates a culture of intolerance and discrimination. This analysis reveals the depth and complexity of hate speech on the platform, illustrating how it is embedded within larger discussions that often include political, social, and cultural dimensions. This understanding underscores the challenges of combating hate speech, as it is not merely a matter of curbing offensive language but also addressing the broader ideological and societal contexts that foster such discourse.

## V. DISCUSSION

This study provides a comprehensive analysis of the prevalence and characteristics of hate speech on the 4chan platform, particularly focusing on the /pol/ board. Utilising advanced NLP techniques, the research quantified and categorised various forms of hate speech.

### A. Discussion of Results

The finding that 11.20% of posts contain hate speech reveals a significant prevalence of harmful content on 4chan, which is especially notable given the platform's anonymity and minimal moderation. This proportion raises concerns about the platform's role in facilitating and perpetuating hate speech. The evident contrast between the hate and non-hate categories accentuates the platform's dual nature as a space for free expression and a breeding ground for toxic discourse. These insights align with the existing literature [1], providing empirical evidence that can inform theoretical discussions on the impact of platform governance on user behaviour.

The detailed breakdown of hate speech into specific categories (e.g., racism, religion, and sexual orientation) quantifies the types of hate prevalent on 4chan and highlights the targeted nature of this hate. Racism being the most common form of hate speech mirrors broader societal issues and points to deep-seated prejudices that transcend the digital realm. The significant presence of hate speech targeting religious groups and individuals based on sexual orientation also highlights the intersectionality of hate. This aligns with previous studies on other platforms [8], [10]. These insights are crucial for developing targeted interventions that address the specific needs of affected communities.

The high incidence of toxicity across hate speech discussions underscores the aggressive and hostile environment on 4chan. The prevalence of identity attacks and the use of derogatory language in these contexts might reflect tactics employed by users to harm others. This is particularly important for content moderators and platform developers, as it calls for advanced detection algorithms that can recognise and mitigate such nuanced forms of abuse. This finding extends previous work on toxic online behaviour [3] by providing a more granular understanding of how toxicity manifests in different types of hate speech.

The application of LDA for topic modelling has uncovered prevalent themes and subtopics within each category of hate speech, offering a window into the complex ways in which hate speech is contextualised on 4chan. The recurrent use of highly offensive language within these discussions not only highlights the aggressive nature of discourse but also sheds light on the cultural and social underpinnings of such behaviour. This aligns with research on topic analysis showing that discussions in /pol/ feature political matters, hate, misogyny, and racism in unmoderated platforms [19].

Our work cannot be directly compared to those listed in Table I, since such studies do not quantify or measure the prevalence of online hate. Unlike our study, which reports specific measurements of online hate, most studies focus on exploring and comparing NLP techniques, datasets, and models in terms of their accuracy or precision. These studies primarily aim to improve the detection capabilities of various models rather than assessing the real-world prevalence of online hate speech.

*B. Research Challenges*

The large-scale dataset that was systematically curated and analysed in this study reflects the broad scope of interactions on 4chan, particularly on the /pol/ board. The removal of 23,241 empty posts underscores the challenges associated with data quality in unmoderated environments such as 4chan. This substantial preprocessing step is vital for ensuring the accuracy of subsequent analyses and highlights the need for robust data-cleaning methods in research involving digital platforms characterised by user-generated content. The extensive volume of valid posts provides a solid foundation for the analysis, lending credibility to the study's findings and facilitating a reliable exploration of hate speech dynamics.

Another challenge was the absence of prior work focused on analysing online hate against different vulnerable groups specifically on 4chan. While there has been research on more mainstream platforms, 4chan remains relatively unexplored. This gap posed unique challenges, as existing methodologies and models had to be adapted or entirely rethought to suit the dynamics of 4chan's /pol/ board.

*C. Takeaways for Stakeholders*

Hate speech in its online form creates complex risks and harms which are sometimes difficult to identify and address. Different stakeholders are involved and responsible for such a phenomenon. The Alan Turing Institute's hub for online hate research (turing.ac.uk/research/research-programmes/public-policy/online-hate-research-hub) provides an extensive collection of research and evidence in this area. The findings of our study can offer insight to various stakeholders as we discuss here.

**Researchers:** While there is a body of research on measuring and tackling online hate, its growth and the complicated ways of its spread call for united forces across different disciplines. This study provides a methodological framework for analysing hate speech in less moderated online environments. Future research could build on this approach to conduct comparative studies across different platforms or to explore longitudinal trends in online hate speech.

**Policymakers:** The widespread presence of hate speech on 4chan highlights the need for more carefully crafted policies that tackle the specific challenges of anonymous, minimally moderated platforms. The minimal moderation on this kind of platforms is frequently justified as a means of safeguarding free speech; however, this study's findings highlight the risks associated with this approach, especially regarding the unchecked proliferation of online hate and toxic content. Striking a balance between necessary moderation and the protection of free speech is a challenging task, as overly strict moderation may suppress legitimate expression, while insufficient oversight can lead to the spread of harmful behaviours [33].

The findings from this study have significant implications in the context of evolving regulatory frameworks such as the UK Online Safety Act and the European Union's online content moderation policies. The UK Online Safety Act (gov.uk/government/publications/online-safety-act-explainer/online-safety-act-explainer), which seeks to impose stringent regulations on online platforms to protect users from harmful content, could directly impact platforms like 4chan, requiring them to implement more robust content moderation strategies to curb the spread of hate speech. Similarly, the EU's regulations (fra.europa.eu/en/publication/2023/online-content-moderation ) on online content moderation emphasise the need for platforms to proactively manage illegal and harmful content, including online hate, with a focus on the protection of fundamental rights.

The EU's efforts to criminalise hate speech under its legal framework (europarl.europa.eu/news/en/press-room/20240112IPR16777/time-to-criminalise-hate-speech-and-hate-crime-under-eu-law ), further highlights the urgency of addressing online hate on platforms like 4chan. The criminalisation of hate speech at the EU level represents a significant shift towards holding individuals and platforms accountable for the spread of harmful content. The findings from this research could inform the implementation of these laws by providing empirical evidence.

**Platform developers:** The detailed analysis of hate speech categories and toxic content can inform the development of more sophisticated content moderation systems. Future efforts could focus on creating context-aware moderation tools that can detect and mitigate nuanced forms of hate speech.

**Society at large:** The findings highlight the urgent need for digital literacy initiatives that educate users about the impacts of online hate speech and promote more responsible online behaviour. More specifically, at-risk groups such as children, older adults, people with disabilities, and people of certain races and genders can be at greater risk. There are several Centres and NGOs such as the Center for Countering Digital Hate (counterhate.com), HOPE NOT HATE (hopenothate.org.uk), and SELMA (hackinghate.eu) who support research, contribute to campaigning, educate communities, and work with policymakers to tackle online hate.

**Others:** Other players like think tanks and private sectors should also contribute to these efforts. In a recent article, the Chatham House (chathamhouse.org/2024/05/gendered-hate-speech-data-breach-and-state-overreach ) discusses three types of cyber harms that have differential risks based on their gender: hate speech, data breach and state overreach (for example, cybercrime laws or other legislation reinforcing discriminatory gender norms online). The intersectional nature of online hate highlighted in our work contributes to these conversations, demonstrating how all sectors should work closely to tackle online hate.

*D. Limitations and Future Work*

Focusing on a single platform limits the generalisability of our findings to other online environments. Future research should explore comparative studies across platforms to better understand online hate speech ecosystems, develop platform-specific models for unique linguistic and cultural contexts, and examine the temporal dynamics of hate speech in relation to real-world events or policy changes.

Although pre-trained NLP models like RoBERTa and Detoxify were effective in this study, they are not fully optimised for 4chan's unique linguistic patterns, such as non-standard language and coded speech on boards like /pol/. Fine-tuning these models with a tailored dataset specific to 4chan would improve accuracy.

## VI. Conclusion

This research has systematically explored the prevalence and characteristics of online hate speech on the platform 4chan, particularly its /pol/ board, employing advanced NLP techniques to analyse a large dataset of posts. The key findings from this study offer valuable insights into the scale and nature of hate speech within an online environment known for its minimal moderation and anonymity. A substantial portion of the posts analysed were identified as containing hate speech, with racism, religious intolerance, and discrimination based on sexual orientation emerging as the most prevalent forms. The implications of this research are significant for the field of online behaviour and platform governance. The detailed analysis of hate speech dynamics on 4chan can inform the development of more sophisticated content moderation technologies that are capable of recognising and mitigating such speech effectively. For policymakers and platform administrators, these findings emphasize the need for targeted interventions that address not only the symptoms but also the underlying cultural and social drivers of online hate.

## BIOGRAPHICAL NOTES

Adrian Bermudez Villalva is a Research Fellow at RHUL. His research focuses on cybersecurity and measuring online malicious activities. Bermudez Villalva received a Ph.D. from UCL, UK.

Maryam Mehrnezhad is a Reader at RHUL. Her research interests are the security and privacy of marginalized and at-risk users. Mehrnezhad received a Ph.D. from Newcastle University, UK.

Ehsan Toreini is a Lecturer at University of Surrey. His research is focused on physical security, trustworthy machine learning and web security. Toreini received a Ph.D from Newcastle University, UK.